\documentclass[english,review,authoryear]{elsarticle}
\usepackage[T1]{fontenc}
\usepackage[latin9]{inputenc}
\usepackage{amsmath}
\usepackage{graphicx}
\usepackage{amssymb}
\usepackage{esint}
\usepackage{babel}
\usepackage{times}
\usepackage{babel}

\numberwithin{equation}{section}

\newtheorem{theorem}{Theorem}[section]
\newtheorem{lemma}[theorem]{Lemma}


\begin{document}
\begin{frontmatter}

\title{An application of the stationary phase method for estimating probability densities of function derivatives}
\author[ufl]{Karthik S. Gurumoorthy\corref{cor1}}
\ead{sgk@ufl.edu}

\author[ufl]{Anand Rangarajan}
\ead{anand@cise.ufl.edu}

\author[ufl]{Arunava Banerjee}
\ead{arunava@cise.ufl.edu}

\cortext[cor1]{Corresponding Author\\
 Address:\\
 E301, CSE Building, University of Florida,\\
 P.O. Box 116120, Gainesville, FL 32611-6120, USA.\\
 Ph: 001-352-392-1200, Fax: 001-352-392-1220 }

\address[ufl]{Department of Computer and Information Science and Engineering,
University of Florida, Gainesville, Florida, USA}

\begin{abstract}
We prove a novel result wherein the density function of the gradients---corresponding to density function of the derivatives in one dimension---of a thrice differentiable function $S$ (obtained via a random variable
transformation of a uniformly distributed random variable) defined
on a closed, bounded interval $\Omega\subset\mathbb{R}$ is accurately
approximated by the normalized power spectrum of $\phi=\exp\left(\frac{iS}{\tau}\right)$
as the free parameter $\tau\rightarrow0$. The result is shown using
the well known stationary phase approximation and standard integration
techniques and requires proper ordering of limits. Experimental results
provide anecdotal visual evidence corroborating the result. 

\end{abstract}
\end{frontmatter}
\section{Introduction}

The literature is replete with techniques which attempt to estimate
a non-observable probability density function using observed data
believed to be sampled from an underlying distribution. Density estimation
techniques have a long history and run the gamut of histogramming,
Parzen windows \cite{Parzen62}, vector quantization, wavelets
etc.  The centerpiece of our current work is to provide a useful application of
the stationary phase method to compute the density function corresponding to the derivatives of the function $S$ (density function of $S^{\prime}$).
Here, the density function of the derivatives is obtained
via a random variable transformation of a uniformly distributed random
variable $X$ (over the bounded interval $\Omega$) using the derivative $S^{\prime}$---denoted by the symbol $s$--- 
as the transformation function. In other words, if we define a random variable $Y=s(X)$
where the random variable $X$ has a \emph{uniform distribution} on the interval $\Omega$, the density
function of $Y$ represents the density function of the derivative $s$. 

We introduce a novel approach for computing the density of $Y$, where we express the given function $S$
as the phase of a wave function $\phi$, specifically $\phi(x)=\exp\left(\frac{iS(x)}{\tau}\right)$
for small values of $\tau$, and then consider the normalized power
spectrum---magnitude squared of the Fourier transform---of $\phi$ 
\cite{Bracewell99}. Using the \emph{stationary phase approximation}---a
well known technique in asymptotic analysis \cite{OlverBook74}---we
show that in the limiting case as $\tau\rightarrow0$, the power spectrum
of $\phi$ converges to the density of $Y$ and hence can serve as
its density estimator at small, non-zero values of $\tau$.
In other words, if $P(u)$ denotes the density of $Y$ and if $P_{\tau}(u)$
corresponds to the power spectrum of $\phi$ at a given value of $\tau$, Theorem~\ref{GradDensityID}
constitutes the following relation, namely 
\begin{equation}
\lim_{\tau\rightarrow0}\int_{u_{0}}^{u_{0}+\alpha}P_{\tau}(u)du=\int_{u_{0}}^{u_{0}+\alpha}P(u) du,
\end{equation}
for a small value of the interval measure $\alpha$ on $u$. We call our approach as the \emph{wave function method} for computing the probability density function and henceforth will refer to it by the same name.
 
We would like to accentuate the following very important points. 
\begin{itemize}
\item Though the mechanism of the stationary phase method makes it a useful tool for the density estimation of function derivatives ($s$),
our wave function approach can easily be extended for determining the densities of a given function rather than its derivative. In other words, a function $s(x)$ can be integrated to obtain $S(x)$ such that $S^{\prime}(x) = s(x)$. Our method constructed on the platform of the stationary phase approximation can then be employed to obtain the density function of $s$.
\item The stationary phase method has little bearing with the characteristic function approach for estimating the probability densities---briefly described in Section~\ref{sec:charfuncformulation} and also see \cite{Billingsley95}---and should be treated as an independent recipe for computing the density function. Though we build an informal bridge between the wave function and the characteristic function approaches in Section~\ref{sec:relationcharfunc}, based on the reasons described therein, we strongly believe that the mechanism of stationary phase approximation is essential to formally prove Theorem~\ref{GradDensityID}. Hence our wave function method should not be merely treated as a reformulation of the characteristic function approach.
\item Our work is \emph{fundamentally different} from estimating the gradients of a density function \cite{Fukunaga75} and should not be semantically confused with it.
\end{itemize}
\subsection{Motivation from quantum mechanics}
Our new mathematical relationship is motivated by a well known relation
in the quantization of classical systems, wherein classical physics is expressed as a limiting case
of quantum mechanics (please refer to \cite{Griffiths04,Feynman65}).
When $S$ is treated as the Hamilton-Jacobi scalar field, the gradient (derivative in one dimension)
of $S$ corresponds to the classical momentum of a particle.
In the parlance of quantum mechanics, the
magnitude square of the wave function expressed either in its position
or momentum basis corresponds to its position or momentum density
respectively. Since these representations (either in the position
or momentum basis) are simply the (suitably scaled) Fourier transforms
of each other, the magnitude square of the Fourier transform of the
wave function expressed in its position basis, is its quantum momentum
density \cite{Griffiths04}. The principal theorem proved
in the article {[}Theorem~\ref{GradDensityID}{]} states that the
classical momentum density (denoted by $P$) can be expressed as a
limiting case (as $\tau\rightarrow0$) of its corresponding quantum
momentum density (denoted by $P_{\tau}$), in complete agreement with
the correspondence principle \cite{Griffiths04}. Our work leverages this relation to provide a new density estimator for the
derivatives.

\section{Existence of the density function}

As stated above, the density function for the derivative $s$ can be obtained via a random variable transformation of a
uniformly distributed random variable $X$ using the derivative $s$
as the transformation function, namely, $Y=s(X)$. We assume
that $S$ is thrice differentiable on a closed, bounded interval $\Omega=[b_{1},b_{2}]$,
with length $L=b_{2}-b_{1}$ and has a non-vanishing second derivative
\emph{almost everywhere} on $\Omega$, i.e. 
\begin{equation}
\mu\left(\{x:S^{\prime\prime}(x)=0\}\right)=0,
\label{eq:muBis0}
\end{equation}
where $\mu$ denotes the Lebesgue measure. The assumption in Equation~\ref{eq:muBis0}
is made in order to ensure that the density function of $Y$ exists
almost everywhere. This is further clarified in Lemma~\ref{densityLemma}
below.

\noindent Define the following sets: 
\begin{eqnarray}
\mathcal{B} & \equiv & \{x:S^{\prime\prime}(x)=s^{\prime}(x)=0\},\nonumber \\
\mathcal{C} & \equiv & \{s(x):x\in\mathcal{B}\}\cup\{s(b_{1}),s(b_{2})\},\,\mathrm{and}\,\nonumber \\
\mathcal{A}_{u} & \equiv & \{x:s(x)=u\}.
\label{AandCset}
\end{eqnarray}
Here, $s(b_{1})=\lim_{x\rightarrow b_{1}^{+}}s(x)$
and $s(b_{2})=\lim_{x\rightarrow b_{2}^{-}}s(x)$.
The higher derivatives of $S$ at the end points $b_{1},b_{2}$ are
also defined along similar lines using one-sided limits. The main purpose of
defining these one-sided limits is to exactly determine the set $\mathcal{C}$ 
where the density of $Y$ is \emph{not defined}.
Since $\mu(\mathcal{B})=0$, we also have $\mu(\mathcal{C})=0$.

\begin{lemma}{[}Finiteness Lemma{]} \label{finitenessLemma} $\mathcal{A}_{u}$
is finite for every $u\notin\mathcal{C}$. \end{lemma}

\begin{lemma}{[}Density Lemma{]} \label{densityLemma} The probability
density of $Y$ on $\mathbb{R}-\mathcal{C}$ exists and is given by
\begin{equation}
P(u_{0})=\frac{1}{L}\sum_{k=1}^{N(u_{0})}\frac{1}{\left|S^{\prime\prime}(x_{k})\right|},
\label{eq:graddensity}
\end{equation}
where the summation is over $\mathcal{A}_{u_{0}}$ (which is the finite
set of locations $x_{k}\in\Omega$ where $s(x_{k})=u_{0}$
as per Lemma~\ref{finitenessLemma}), with $\left|\mathcal{A}_{u_{0}}\right|=N(u_{0})$.
\end{lemma} 
Since the density is based on the transformation $Y=s(X)$,
the probability density function in (\ref{eq:graddensity}) assumes
the existence of the inverse of the transformation function $s$
\cite{Billingsley95}. This is made explicit in the following
lemma which is required by the main theorem.

\begin{lemma}{[}Interval Lemma{]} \label{intervalLemma} For every
$u\notin\mathcal{C}$, $\exists\eta>0$ and a closed interval $\mathcal{J}_{\eta}=[u-\eta,u+\eta]$
such that $\mathcal{J}_{\eta}\cap\mathcal{C}$ is empty. 
\end{lemma}
The proofs of the lemmas are available in \ref{sec:Proof-of-Lemmas}.

\section{Characteristic function formulation for obtaining densities}
\label{sec:charfuncformulation}
The characteristic function $\psi_Y(\omega)$ for the random variable $Y$ is defined as the expected value of $ \exp\left(i\omega s(X)\right)$, namely
\begin{equation}
\psi_Y(\omega) \equiv E\left[ \exp\left(i\omega s(X)\right) \right] = \frac{1}{L} \int_{b_1}^{b_2} \exp\left(i\omega s(x)\right) dx.
\label{eq:characfunc}
\end{equation} 
Note that, here $\frac{1}{L}$ denotes the density of the uniformly distributed random variable $X$ on $\Omega$.

The inverse Fourier transform of characteristic functions also serves as the density functions of random variables \cite{Billingsley95}. In other words, the density
function $P(u)$ of the random variable $Y$ can be obtained via
\begin{equation}
P(u) = \frac{1}{2\pi} \int_{-\infty}^{+\infty}\psi_Y(\omega) \exp(-i\omega u) d\omega = \frac{1}{2\pi L}\int_{-\infty}^{+\infty} \int_{b_1}^{b_2} \exp\left\{i\omega (s(x)-u)\right\} dx d\omega.
\label{eq:iftcharfunc}
\end{equation}

In this work, we also showcase a direct relation between the characteristic function formulation for computing densities and the power spectrum of the wave function $\phi(x) = \exp\left(\frac{iS(x)}{\tau}\right)$ at small values of $\tau$, as elucidated in Section~\ref{sec:relationcharfunc}.

\section{Equivalence of the gradient density and the power spectrum}
We now prove the main result which relates the normalized power spectrum
of $\exp\left(\frac{iS(x)}{\tau}\right)$ (in the limit as $\tau\rightarrow0$)
with the probability density of the random variable $Y=s(X)$
(denoted by $P$).

Define a function $F:\mathbb{R}\times\mathbb{R}^{+}\rightarrow\mathbb{C}$
as \begin{equation}
F(u,\tau)\equiv\frac{1}{\sqrt{2\pi\tau L}}\int_{b_{1}}^{b_{2}}\exp\left(\frac{iS(x)}{\tau}\right)\exp\left(\frac{-iux}{\tau}\right)dx.
\label{eq:Fuhbar}
\end{equation}
 For a fixed value of $\tau$, define a function $F_{\tau}:\mathbb{R}\rightarrow\mathbb{C}$
as \begin{equation}
F_{\tau}(u)\equiv F(u,\tau).\label{eq:Fhbar}\end{equation}
 Observe that $F_{\tau}$ is closely related to the Fourier transform
of $\exp\left(\frac{iS(x)}{\tau}\right)$. The scale factor $\frac{1}{\sqrt{2\pi\tau L}}$
is the normalizing term such that the $L_{2}$ norm of $F_{\tau}$
is one, as seen in the following lemma (whose proof is straightforward
and omitted here).

\begin{lemma} \label{lemma1} With $F_{\tau}$ defined as above,
$F_{\tau}\in L^{2}(\mathbb{R})$ and $\|F_{\tau}\|=1$. \end{lemma}
Define a function $P_{\tau}:\mathbb{R}-\mathcal{C}\rightarrow\mathbb{R}^{+}$
as \begin{equation}
P_{\tau}(u)\equiv\left|F_{\tau}(u)\right|^{2}=F_{\tau}(u)\overline{F_{\tau}(u)}.\label{def:Ph}\end{equation}
 By definition, $P_{\tau}\geq0$. Since $\mu(\mathcal{C})=0$, from
Lemma~(\ref{lemma1}), $\int_{-\infty}^{\infty}P_{\tau}(u)du=1$.
Hence, treating $P_{\tau}(u)$ as a density function, we have the
following theorem statement.

\begin{theorem} \label{GradDensityID} If $P$ and $P_{\tau}$ are
defined as above, then the \begin{equation}
\lim_{\alpha\rightarrow0}\frac{1}{\alpha}\lim_{\tau\rightarrow0}\int_{u_{0}}^{u_{0}+\alpha}P_{\tau}(u)du=P(u_{0}),\hspace{2pt}\forall u_{0}\notin\mathcal{C}.
\end{equation}
 \end{theorem} 
 Before embarking on the proof, we would like to emphasize
that the ordering of the limits and the integral as given in the theorem
statement is crucial and cannot be arbitrarily interchanged. To press
this point home, we show below that after solving for $P_{\tau}$,
the $\lim_{\tau\rightarrow0}P_{\tau}$ does not exist. Hence, the
order of the integral followed by the limit $\tau\rightarrow0$ cannot
be interchanged. Furthermore, when we swap the limits between $\alpha$
and $\tau$, we get \begin{equation}
\lim_{\tau\rightarrow0}\lim_{\alpha\rightarrow0}\frac{1}{\alpha}\int_{u_{0}}^{u_{0}+\alpha}P_{\tau}(u)du=\lim_{\tau\rightarrow0}P_{\tau}(u_{0})\end{equation}
which does not exist. Hence, the theorem statement is valid \emph{only}
for the specified sequence of limits and the integral.

\subsection{Brief exposition of the result}

\noindent To understand the result in simpler terms, let us reconsider
the definition of the scaled Fourier transform given in Equation~\ref{eq:Fuhbar}.
The first exponential $\exp\left(\frac{iS}{\tau}\right)$ is a varying
complex {}``sinusoid\textquotedblright{}, whereas the second exponential
$\exp\left(\frac{-iux}{\tau}\right)$ is a fixed complex sinusoid
at frequency $\frac{u}{\tau}$. When we multiply these two complex
exponentials, at low values of $\tau$, the two sinusoids are usually
not {}``in sync\textquotedblright{} and tend to cancel each other
out. However, around the locations where $s(x)=u$, the two
sinusoids are in perfect sync (as the combined exponent is \emph{stationary})
with the approximate duration of this resonance depending on $S^{\prime\prime}(x)$.
The value of the integral in (\ref{eq:Fuhbar}) can be approximated
via the stationary phase approximation \cite{OlverBook74}
as \begin{equation}
F_{\tau}(u)\approx\frac{1}{\sqrt{L}}\exp\left(\pm i\frac{\pi}{4}\right)\sum_{k=1}^{N(u)}\exp\left\{ \frac{i}{\tau}\left(S(x_{k})-ux_{k}\right)\right\} \frac{1}{\sqrt{S^{\prime\prime}(x_{k})}}\end{equation}
where $N(u)=\left|\mathcal{A}_{u}\right|$. The approximation is increasingly
tight as $\tau\rightarrow0$. The squared Fourier transform ($P_{\tau}$)
gives us the required result $\frac{1}{L}\sum_{k=1}^{N(u)}\frac{1}{\left|S^{\prime\prime}(x_{k})\right|}$
except for the \emph{cross} phase factors $S(x_{k})-S(x_{l})-u(x_{k}-x_{l})$
obtained as a byproduct of two or more remote locations $x_{k}$ and
$x_{l}$ indexing into the same frequency bin $u$, i.e, $x_{k}\not=x_{l}$,
but $s(x_{k})=s(x_{l})=u$. Integrating the squared
Fourier transform over a small frequency range $[u,u+\alpha]$ removes
these cross phase factors and we obtain the desired result.

\subsection{Relation to characteristic function formulation}
\label{sec:relationcharfunc}
Before we demonstrate the relation between the power spectrum of $\phi$, namely $P_{\tau}$, as $\tau \rightarrow 0$ and the inverse Fourier transform of the characteristic function $\psi_Y$ as given in Equation~\ref{eq:iftcharfunc}, we would like to accentuate the following precautionary note. What we show below \emph{cannot} be treated as a formal proof of Theorem~\ref{GradDensityID}.  The main motivation of this section is to provide an intuitive reason behind our theorem, where we try to directly manipulate the power spectrum of $\phi$ into the characteristic function formulation given in Equation~\ref{eq:iftcharfunc}, circumventing the need for the closed-form expression of the density function $P(u)$ given in Equation~\ref{eq:graddensity}. Though we do not obtain a formal proof through this approach, our attempt helps to showcase the profound connection between the power spectrum and the characteristic function formulation and thereby to the density function $P(u)$. The formal proof is given in the subsequent section by using the method of stationary phase approximation.

Based on the expression for the scaled Fourier transform $F_{\tau}(u)$ in Equation~\ref{eq:Fuhbar}, the power spectrum $P_{\tau}(u)$ is given by
\begin{equation}
P_{\tau}(u) = \frac{1}{2 \pi \tau L}\int_{b_1}^{b_2} \int_{b_1}^{b_2} \exp\left(\frac{i}{\tau}[S(x)-S(y)]\right) \exp\left(\frac{-i u}{\tau}[x-y]\right) dx dy.
\end{equation}
Define the following change of variables namely, $\omega = \frac{x-y}{\tau}$ and $\nu=\frac{x+y}{2}$. Then we get 
\begin{equation}
x = \nu + \frac{\omega \tau}{2}, y = \nu - \frac{\omega \tau}{2}, dx dy = \tau d\omega d\nu,
\end{equation}
and $P_{\tau}$ can be rewritten as
\begin{equation}
P_{\tau}(u) = \frac{1}{2 \pi L} \left(P_{\tau}^{(1)}(u) + P_{\tau}^{(2)}(u) \right)
\end{equation}
where,
\begin{equation}
P_{\tau}^{(1)}(u) = \int_{\frac{b_1-b_2}{\tau}}^{0} \xi(\omega,u) d\omega, \hspace{10pt} P_{\tau}^{(2)}(u) = \int_{0}^{\frac{b_2-b_1}{\tau}}  \xi(\omega,u) d\omega.
\end{equation}
Here $\xi(\omega,u)$ denotes the integral over $\nu$ and is given by
\begin{equation}
\xi(\omega,u) = \int_{b_1+\frac{|\omega| \tau}{2}}^{b_2-\frac{|\omega| \tau}{2}} \exp\left(\frac{i}{\tau}\left[S\left(\nu+\frac{\omega \tau}{2}\right)-S\left(\nu - \frac{\omega \tau}{2}\right)\right]\right) \exp(-i u\omega) d\nu.
\end{equation}
By applying the mean value theorem, we can write
\begin{equation}
\label{eq:meanvaluethm}
S\left(\nu+\frac{\omega \tau}{2}\right)-S\left(\nu - \frac{\omega \tau}{2}\right) = s(z) \omega \tau,
\end{equation}
where $z \in \left(\nu - \frac{\omega \tau}{2}, \nu + \frac{\omega \tau}{2}\right)$. This allows $\xi(\omega,u)$ to be reformulated as
\begin{equation}
\label{eq:xi}
\xi(\omega,u) = \int_{b_1+\frac{|\omega| \tau}{2}}^{b_2-\frac{|\omega| \tau}{2}} \exp\left\{i \omega s\left(z\left(\nu,\frac{\omega\tau}{2}\right)\right)\right\} \exp(-i u\omega) d\nu.
\end{equation}
Note that for a fixed value of $\omega$, $\omega \tau \rightarrow 0$ as $\tau \rightarrow 0$. Hence for a fixed value of $\nu$ and $\omega$, $z \rightarrow \nu$ as $\tau \rightarrow 0$. So for small values of $\tau$,
we can approximate $\xi(\omega,u)$ by
\begin{equation}
\label{eq:xiapprox}
\xi(\omega,u) \approx \int_{b_1}^{b_2} \exp(i \omega s(\nu)) \exp(-i u\omega) d\nu.
\end{equation}
Again we would like to drive the following point home. We \emph{do not} claim that we have formally proven the above approximation. On the contrary, we believe that it might be an onerous task to do so as the mean value theorem point $z$ is unknown and the integration limits for $\nu$ directly depend on $\tau$. The approximation is stated with the sole purpose of providing an intuitive reason for our Theorem~\ref{GradDensityID}.

Observe that the integral range for $\omega$ in $P_{\tau}^{(1)}(u)$ (and also in $P_{\tau}^{(2)}(u)$) is a function of $\tau$. So when $\omega = O\left(\frac{1}{\tau}\right)$, $\omega \tau \not\rightarrow 0$ as $\tau \rightarrow 0$ and hence the above approximation for $\xi(\omega,u)$ in Equation~\ref{eq:xiapprox} might seem to break down. To elude this problem, we divide the integral range for $\omega$ in $P_{\tau}^{(1)}(u)$ into two disjoint regions. To this end, consider a small $\epsilon >0$ and let $M = (b_2-b_1)\tau^{\epsilon-1}$. Then $P_{\tau}^{(1)}(u)$ can be split into two integrals namely,
\begin{equation}
\label{eq:Ptau1split}
P_{\tau}^{(1)}(u) = \int_{\frac{b_1-b_2}{\tau}}^{-M} \xi(\omega,u) d\omega + \int_{-M}^{0} \xi(\omega,u) d\omega.
\end{equation}
In the second integral of Equation~\ref{eq:Ptau1split}, $\omega$ is deliberately made to be $O(\tau^{\epsilon-1})$ and hence $\omega \tau \rightarrow 0$ as $\tau \rightarrow 0$. Hence the approximation for $\xi(\omega,u)$ in Equation~\ref{eq:xiapprox} might hold for this integral range of $\omega$. Now consider the first integral of Equation~\ref{eq:Ptau1split} and recall that the Theorem~\ref{GradDensityID} requires the power spectrum $P_{\tau}(u)$ to be integrated over a small region $[u_0, u_0+\alpha]$. By using the true expression for $\xi(\omega,u)$ from Equation~\ref{eq:xi} and performing the integral for $u$ prior to $\omega$ and $\nu$, we get
\begin{align}
& \int_{u_0}^{u_0+\alpha}\int_{\frac{b_1-b_2}{\tau}}^{-M} \xi(\omega,u) d\omega du = \nonumber \\
& \int_{\frac{b_1-b_2}{\tau}}^{-M}\int_{b_1+\frac{|\omega| \tau}{2}}^{b_2-\frac{|\omega| \tau}{2}} \exp\left\{i \omega s\left(z\left(\nu,\frac{\omega\tau}{2}\right)\right)\right\} \int_{u_0}^{u_0+\alpha} \exp(-i u\omega) du d\nu d\omega.
\end{align}
Since both $M$ and $\frac{b_2-b_1}{\tau}$ approach $\infty$ as $\tau \rightarrow0$, from the Riemann-Lebesgue lemma we see that $\forall \omega \in \left[\frac{b_1-b_2}{\tau}, -M \right]$, the integral
$\int_{u_0}^{u_0+\alpha} \exp(-i u\omega) du \rightarrow 0$ as $\tau \rightarrow 0$. Hence for small values of $\tau$, we can expect the first integral in Equation~\ref{eq:Ptau1split} to dominate over the second integral. This leads to the following approximation namely,
\begin{equation}
\label{eq:Ptau1approx}
\int_{u_0}^{u_0+\alpha} P_{\tau}^{(1)}(u)  du \approx \int_{u_0}^{u_0+\alpha} \int_{-M}^{0} \xi(\omega,u) d\omega du,
\end{equation}
as $\tau$ approaches zero. A similar approximation can also be obtained for the integral of $P_{\tau}^{(2)}(u)$ over $[u_0, u_0+\alpha]$ at small values of $\tau$. Combining the above approximation with the approximation for $\xi(\omega,u)$ given in Equation~\ref{eq:xiapprox} and noting that $M \rightarrow \infty$ as $\tau \rightarrow 0$, the integral of the power spectrum $P_\tau(u)$ over the interval $[u_0, u_0+\alpha]$ at small values of $\tau$, can be approximated by
\begin{equation}
\int_{u_0}^{u_0+\alpha} P_{\tau}(u)  du \approx \frac{1}{2 \pi L} \int_{u_0}^{u_0+\alpha} \int_{-\infty}^{\infty} \int_{b_1}^{b_2} \exp\left\{i \omega \left(s(\nu)- u\right)\right\} d\nu d\omega du.
\end{equation}
This form \emph{exactly} coincides with the expression given in Equation~\ref{eq:iftcharfunc} obtained through the characteristic function formulation.

We believe that the approximations given in Equations~\ref{eq:xiapprox} and ~\ref{eq:Ptau1approx} cannot be proven easily as they involve limits of integration which directly depend on $\tau$. Furthermore, the mean value theorem point $z$ in Equation~\ref{eq:meanvaluethm} is arbitrary and cannot be determined beforehand for a given value of $\tau$. Nevertheless, we propose to take a different path and use the method of stationary phase approximation to formally prove Theorem~\ref{GradDensityID}.

\subsection{Formal proof}

We shall now provide the proof by considering different cases.

\noindent \textbf{case (i)}: Let us consider the case in which no
stationary points exist for the given $u_{0}$, i.e, there is no $x\in\Omega$
for which $s(x)=u_{0}$. Let $t(x)=S(x)-u_{0}x$. Then, $t'(x)$
is of constant sign in $[b_{1},b_{2}]$ and hence $t(x)$ is strictly
monotonic. Defining $v=t(x)$, we have from Equation~\ref{eq:Fuhbar}, 
\begin{equation}
F_{\tau}(u_{0})=\frac{1}{\sqrt{2\pi\tau L}}\int_{t(b_{1})}^{t(b_{2})}\exp\left(\frac{iv}{\tau}\right)g(v)dv.
\end{equation}
 Here, $g(v)=\frac{1}{t'(x)}$ where $x=t^{-1}(v)$. Integrating by
parts, we get 
\begin{eqnarray}
F_{\tau}(u_{0})\sqrt{2\pi\tau L} & = & \frac{\tau}{i}\left[\exp\left(\frac{it(b_{2})}{\tau}\right)g\left(t(b_{2})\right)-\exp\left(\frac{it(b_{1})}{\tau}\right)g\left(t(b_{1})\right)\right]\nonumber \\
 &  & -\frac{\tau}{i}\int_{t(b_{1})}^{t(b_{2})}\exp\left(\frac{iv}{\tau}\right)g^{\prime}(v)dv.
 \end{eqnarray}
Then 
 \begin{equation}
\left|F_{\tau}(u_{0})\right|\sqrt{2\pi\tau L}\leq\tau\left(\frac{1}{|s(b_{2})-u_{0}|}+\frac{1}{|s(b_{1})-u_{0}|}+\int_{t(b_{1})}^{t(b_{2})}\left|g^{\prime}(v)\right|dv\right).
\end{equation}
 Hence, $\left|F_{\tau}(u_{0})\right|\leq\gamma_{1}(u_{0})\sqrt{\tau}$,
where $\gamma_{1}(u_{0})>0$ is a continuous function of $u_{0}$.
Then $P_{\tau}(u_{0})\leq\gamma_{1}^{2}(u_{0})\tau$. Since $s(x)$
is continuous and $u_{0}\notin\mathcal{C}$, we can find a $\rho>0$
such that for every $u\in[u_{0}-\rho,u_{0}+\rho]$, no stationary
points exist. The value $\rho$ can also be chosen appropriately such
that $[u_{0}-\rho,u_{0}+\rho]\cap\mathcal{C}=\emptyset$. If $|\alpha|<\rho$,
then 
\begin{equation}
\lim_{\tau\to0}\int_{u_{0}}^{u_{0}+\alpha}P_{\tau}(u)du=0.
\end{equation}
 Furthermore, from Equation~\ref{eq:graddensity} we have $P(u_{0})=0$.
The result immediately follows.\\

\noindent \textbf{case (ii)}: We now consider the case where stationary
points exist. Since we are interested only in the situation as $\tau\rightarrow0$,
the stationary phase method in \cite{OlverBook74,OlverArticle74}
can be used to obtain a good approximation for $F_{\tau}(u_{0})$
defined in Equations~\ref{eq:Fuhbar} and \ref{eq:Fhbar}. The phase term
in this function, $\frac{S(x)-u_{0}x}{\tau}$, is stationary only
when $s(x)=u_{0}$. Consider the set $\mathcal{A}_{u_{0}}$ defined
in Equation~\ref{AandCset}. Since it is finite by Lemma~\ref{finitenessLemma},
let $\mathcal{A}_{u_{0}}=\{x_{1},x_{2},\ldots,x_{N(u_{0})}\}$ with
$x_{k}<x_{k+1},\forall k$. We break $\Omega$ into disjoint intervals
such that each interval has utmost one stationary point. To this end,
consider numbers $\{c_{1},c_{2},\ldots,c_{N(u_{0})+1}\}$ such that
$b_{1}<c_{1}<x_{1}$, $x_{k}<c_{k+1}<x_{k+1}$ and $x_{N(u_{0})}<c_{N(u_{0})+1}<b_{2}$.
Let $t(x)=S(x)-u_{0}x$. Then, 
\begin{equation}
F_{\tau}(u_{0})\sqrt{2\pi\tau L}=G_{1}+G_{2}+\sum_{k=1}^{N(u_{0})}(K_{k}+\tilde{K}_{k})
\label{eq:FhbarInt}
\end{equation}
where, 
\begin{eqnarray}
G_{1} & = & \int_{b_{1}}^{c_{1}}\exp\left(\frac{it(x)}{\tau}\right)dx,\nonumber \\
G_{2} & = & \int_{c_{N(u_{0})+1}}^{b_{2}}\exp\left(\frac{it(x)}{\tau}\right)dx,\nonumber \\
K_{k} & = & \int_{c_{k}}^{x_{k}}\exp\left(\frac{it(x)}{\tau}\right)dx,\,\mathrm{and}\,\nonumber \\
\tilde{K}_{k} & = & \int_{x_{k}}^{c_{k+1}}\exp\left(\frac{it(x)}{\tau}\right)dx.
\end{eqnarray}
Note that the integrals $G_{1}$ and $G_{2}$ do not have any stationary
points. From case (i) above, we get 
\begin{equation}
G_{1}+G_{2}=\epsilon_{1}(u_{0},\tau)=O(\tau)\label{L1L2bound}
\end{equation}
 as $\tau\rightarrow0$. Furthermore, $\epsilon_{1}(u_{0},\tau)\leq\gamma_{2}(u_{0})\tau$,
where $\gamma_{2}(u_{0})>0$ is a continuous function of $u_{0}$.
In order to evaluate $K_{k}$ and $\tilde{K}_{k}$, observe that when
we expand the phase term up to second order, $t(x)-t(x_{k})\rightarrow Q(x_{k})(x-x_{k})^{2}$
as $x\rightarrow x_{k}$, where $Q(x_{k})=\frac{S^{\prime\prime}(x_{k})}{2}$.
Furthermore, in the open intervals $(c_{k},x_{k})$ and $(x_{k},c_{k+1})$,
$t'(x)=s(x)-u_{0}$ is continuous and is of constant sign.
From Theorem~13.1 in \cite{OlverBook74}, we get 
\begin{equation}
\tilde{K}_{k}=K_{k}=\frac{1}{2}\exp\left(\pm\frac{i\pi}{4}\right)\Gamma\left(\frac{1}{2}\right)\exp\left(\frac{it(x_{k})}{\tau}\right)\sqrt{\frac{2\tau}{\left|S^{\prime\prime}(x_{k})\right|}}+\epsilon_{2}(u_{0},\tau).\label{eq:Ktildek}
\end{equation}
 From Lemma~12.3 in \cite{OlverBook74}, it can be verified that
$\epsilon_{2}(u_{0},\tau)=o(\sqrt{\tau})$ as $\tau\rightarrow0$
and can also be \emph{uniformly bounded} by a function of $u_{0}$
(independent of $\tau$) for small values of $\tau$. In Equation~\ref{eq:Ktildek},
$\Gamma$ is the Gamma function and the sign in the phase term depends
on whether $S^{\prime\prime}(x_{k})>0$ or $S^{\prime\prime}(x_{k})<0$.
Plugging the values of these integrals in Equation~\ref{eq:FhbarInt} and
noting that $\Gamma\left(\frac{1}{2}\right)=\sqrt{\pi}$, we get 
\begin{eqnarray}
F_{\tau}(u_{0})\sqrt{2\pi\tau L} & = & \sum_{k=1}^{N(u_{0})}\exp\left(\frac{i}{\tau}\left[S(x_{k})-u_{0}x_{k}\right]\right)\sqrt{\frac{2\pi\tau}{\left|S^{\prime\prime}(x_{k})\right|}}\exp\left(\pm\frac{i\pi}{4}\right)\nonumber \\
 &  & +\epsilon_{1}(u_{0},\tau)+\epsilon_{2}(u_{0},\tau).
 \end{eqnarray}
 Hence, 
 \begin{eqnarray}
F_{\tau}(u_{0}) & = & \frac{1}{\sqrt{L}}\sum_{k=1}^{N(u_{0})}\frac{\exp\left(\frac{i}{\tau}\left[S(x_{k})-u_{0}x_{k}\right]\right)}{\sqrt{\left|S^{\prime\prime}(x_{k})\right|}}\exp\left(\pm\frac{i\pi}{4}\right)\label{eq:Ftau_approx}\\
 &  & +\epsilon_{3}(u_{0},\tau),
 \end{eqnarray}
 where $\epsilon_{3}(u_{0},\tau)=\frac{\epsilon_{1}(u_{0},\tau)+\epsilon_{2}(u_{0},\tau)}{\sqrt{2\pi\tau L}}=o(1)$
as $\tau\rightarrow0$.

From the definition of $P_{\tau}(u)$ in Equation~\ref{def:Ph}, we have
\begin{eqnarray}
P_{\tau}(u_{0}) & = & \frac{1}{L}\sum_{k=1}^{N(u_{0})}\frac{1}{\left|S^{\prime\prime}(x_{k})\right|}\nonumber \\
 &  & +\frac{1}{L}\sum_{k=1}^{N(u_{0})}\sum_{l=1;l\not=k}^{N(u_{0})}\frac{\cos\left(\frac{1}{\tau}\left[S(x_{k})-S(x_{l})-u_{0}(x_{k}-x_{l})\right]+\theta(x_{k},x_{l})\right)}{\left|S^{\prime\prime}(x_{k})\right|^{\frac{1}{2}}\left|S^{\prime\prime}(x_{l})\right|^{\frac{1}{2}}}\nonumber \\
 &  & +\epsilon_{4}(u_{0},\tau),\label{eq:Ph}
 \end{eqnarray}
 where $\epsilon_{4}(u_{0},\tau)$ includes both the magnitude square
of $\epsilon_{3}(u_{0},\tau)$ and the cross terms involving the main
(first) term in Equation~\ref{eq:Ftau_approx} and $\epsilon_{3}(u_{0},\tau)$.
Notice that the main term in Equation~\ref{eq:Ftau_approx} can be bounded
by a function of $u_{0}$ \emph{independent} of $\tau$ as 
\begin{equation}
\left|\exp\left(\frac{i}{\tau}\left[S(x_{k})-u_{0}x_{k}\right]\right)\right|=1,\forall\tau
\end{equation}
 and $S^{\prime\prime}(x_{k})\not=0,\forall k$. Since $\epsilon_{3}(u_{0},\tau)=o(1)$,
we get $\epsilon_{4}(u_{0},\tau)=o(1)$ as $\tau\rightarrow0$. Additionally,
$\theta(x_{k},x_{l})=0,\,\frac{\pi}{2}$ (or) $-\frac{\pi}{2}$ and
$\theta(x_{l},x_{k})=-\theta(x_{k},x_{l})$.

The first term in Equation~\ref{eq:Ph} exactly matches the expression for
$P(u_{0})$ as seen from Lemma~\ref{densityLemma}. But, since
$\lim_{\tau\rightarrow0}\cos\left(\frac{1}{\tau}\right)$ is not defined,
$\lim_{\tau\rightarrow0}P_{\tau}(u_{0})$ does not exist and hence
the cross cosine terms do not vanish when we take the limit. We now
show that integrating $P_{\tau}(u)$ over a small non-zero interval
$[u_{0},u_{0}+\alpha]$ and then taking the limit with respect to
$\tau$ (followed by the limit with respect to $\alpha)$ does yield
the density of $Y$.

From Lemmas~\ref{finitenessLemma} and \ref{intervalLemma},
we see that for a given $a\in\mathcal{A}_{u_{0}}$, when $u$ is varied
over the interval $\mathcal{J}_{\eta}=[u_{0}-\eta,u_{0}+\eta]$, the
inverse function $\left(s\right)^{-1}(u)$ is well defined
with $\left(s\right)^{-1}(u)\in\mathcal{N}_{a}$, where $\mathcal{N}_{a}$
is some small neighborhood around $a$. For each $a\in\mathcal{A}_{u_{0}}$,
define the inverse function $\left(s_{a}\right)^{-1}(u):\mathcal{J}_{\eta}\rightarrow\mathcal{N}_{a}$
as 
\begin{equation}
\left(s_{a}\right)^{-1}(u)=x\mbox{\,\ iff\,}u=s(x)\mbox{ and }x\in\mathcal{N}_{a}.\label{def:invfuncdef}
\end{equation}
 Unfortunately, when we move from a fixed value $u_{0}$ to a variable
$u$ defined in the interval $\mathcal{J}_{\eta}$, the locations
$x_{k}$ and $x_{l}$ which previously were fixed in Equation~\ref{eq:Ph}
now also vary over the interval $\mathcal{J}_{\eta}.$ This makes
the notation somewhat cumbersome and unwieldy. Using the inverse functions
defined in Equation~\ref{def:invfuncdef} and defining $a_{k}\equiv x_{k}(u_{0})$
{[}and consequently $a_{l}=x_{l}(u_{0})${]}, we define $x_{k}(u)\equiv\left(s_{a_{k}}\right)^{-1}(u)$
{[}and therefore $x_{l}(u)=\left(s_{a_{l}}\right)^{-1}(u)${]}
for $u\in\mathcal{J}_{\eta}$. Finally, define the functions $p_{kl}(u)$
and $q_{kl}(u)$ over $\mathcal{J}_{\eta}$ as 
\begin{eqnarray}
p_{kl}(u) & \equiv & S\left(x_{k}(u)\right)-S\left(x_{l}(u)\right)-u\left(x_{k}(u)-x_{l}(u)\right),\,\mathrm{and}\,\\
q_{kl}(u) & \equiv & \frac{1}{\left|S^{\prime\prime}\left(x_{k}(u)\right)\right|{}^{\frac{1}{2}}\left|S^{\prime\prime}\left(x_{l}(u)\right)\right|^{\frac{1}{2}}}.
\end{eqnarray}
 Observe that 
 \begin{eqnarray}
p_{kl}^{\prime}(u) & = & s(x_{k}(u))x_{k}^{\prime}(u)-s(x_{l}(u))x_{l}^{\prime}(u)-(x_{k}(u)-x_{l}(u))-u\left(x_{k}^{\prime}(u)-x_{l}^{\prime}(u)\right)\nonumber \\
 & = & x_{l}(u)-x_{k}(u)
 \end{eqnarray}
 as $u=s(x_{k}(u))=s(x_{l}(u))$. In particular,
if $x_{l}(u_{0})>x_{k}(u_{0})$, then $x_{l}(u)>x_{k}(u)$ and vice
versa. Hence, $p_{kl}^{\prime}(u)$ never vanishes and is also of
constant sign over $\mathcal{J}_{\eta}$. Then, it follows that $p_{kl}(u)$
is \emph{strictly monotonic} and specifically bijective on $\mathcal{J}_{\eta}$.
We will use this result in the subsequent steps.

Now, let $|\alpha|<\eta$. Since the additional error term $\epsilon_{4}(u_{0},\tau)$
in Equation~\ref{eq:Ph} converges to zero as $\tau\rightarrow0$ and can
also be \emph{uniformly bounded} by a function of $u_{0}$ for small
values of $\tau$, we have 
\begin{equation}
\lim_{\tau\rightarrow0}\int_{u_{0}}^{u_{0}+\alpha}\epsilon_{4}(u_{0},\tau)du=0.
\end{equation}
 Then we get 
 \begin{equation}
\lim_{\tau\rightarrow0}\int_{u_{0}}^{u_{0}+\alpha}P_{\tau}(u)du=I_{1}+I_{2}\label{eq:approxIntP2}
\end{equation}
 where 
 \begin{eqnarray}
I_{1} & \equiv & \frac{1}{L}\sum_{k=1}^{N(u_{0})}\int_{u_{0}}^{u_{0}+\alpha}\frac{1}{\left|S^{\prime\prime}\left(x_{k}(u)\right)\right|}du,\,\mathrm{and}\,\label{eq:I1}\\
I_{2} & \equiv & \frac{1}{L}\sum_{k=1}^{N(u_{0})}\sum_{l=1;l\not=k}^{N(u_{0})}\lim_{\tau\rightarrow0}I_{3}(k,l).\label{eq:I2}
\end{eqnarray}
 Here, $I_{3}(k,l)$ is given by 
 \begin{equation}
I_{3}(k,l)\equiv\int_{u_{0}}^{u_{0}+\alpha}q_{kl}(u)\cos\left[\frac{p_{kl}(u)}{\tau}+\theta\left(x_{k}(u),x_{l}(u)\right)\right]du.\label{eq:I3}
\end{equation}

When $|\alpha|<\eta$, the sign of $S^{\prime\prime}\left(x_{k}(u)\right)$
around $x_{k}(u_{0})$ and the sign of $S^{\prime\prime}\left(x_{l}(u)\right)$
around $x_{l}(u_{0})$ do not change over the interval $[u_{0}$,$u_{0}+\alpha]$.
Since $\theta(x_{k}(u),x_{l}(u))$ depends on the sign of $S^{\prime\prime}$,
$\theta$ is constant on $[u_{0}$,$u_{0}+\alpha]$ and equals $\theta_{kl}=\theta(x_{k}(u_{0}),x_{l}(u_{0}))$.
Now expanding the cosine term in Equation~\ref{eq:I3}, we get 
\begin{eqnarray}
I_{3}(k,l) & = & \cos(\theta_{kl})\int_{u_{0}}^{u_{0}+\alpha}q_{kl}(u)\cos\left(\frac{p_{kl}(u)}{\tau}\right)du\nonumber \\
 &  & -\sin(\theta_{kl})\int_{u_{0}}^{u_{0}+\alpha}q_{kl}(u)\sin\left(\frac{p_{kl}(u)}{\tau}\right)du.
 \end{eqnarray}
Since $p_{kl}(u)$ is bijective, we get via a standard change of
variables: 
\begin{equation}
I_{3}(k,l)=\cos(\theta_{kl})I_{4}(k,l)-\sin(\theta_{kl})I_{5}(k,l)\label{eq:I3equality}
\end{equation}
where 
\begin{eqnarray}
I_{4}(k,l) & = & \int_{\beta_{kl}^{(1)}}^{\beta_{kl}^{(2)}}\cos\left(\frac{v}{\tau}\right)g_{kl}(v)dv,\,\mathrm{and}\\
I_{5}(k,l) & = & \int_{\beta_{kl}^{(1)}}^{\beta_{kl}^{(2)}}\sin\left(\frac{v}{\tau}\right)g_{kl}(v)dv.
\end{eqnarray}
 Here, $\beta_{kl}^{(1)}=p_{kl}(u_{0}),\,\beta_{kl}^{(2)}=p_{kl}(u_{0}+\alpha)$
and $g_{kl}(v)=\frac{q_{kl}(p_{kl}^{-1}(v))}{p_{kl}^{\prime}(p_{kl}^{-1}(v))}$.

Integrating $I_{4}(k,l)$ by parts, we get 
\begin{eqnarray}
I_{4}(k,l) & = & \tau\sin\left(\frac{\beta_{kl}^{(2)}}{\tau}\right)g_{kl}\left(\beta_{kl}^{(2)}\right)-\tau\sin\left(\frac{\beta_{kl}^{(1)}}{\tau}\right)g_{kl}\left(\beta_{kl}^{(1)}\right)\nonumber \\
 &  & -\tau\int_{\beta_{kl}^{(1)}}^{\beta_{kl}^{(2)}}\sin\left(\frac{v}{\tau}\right)g_{kl}^{\prime}(v)dv.
\end{eqnarray}
Then, 
\begin{equation}
\left|I_{4}(k,l)\right|\leq\tau\left(g_{kl}\left(\beta_{kl}^{(2)}\right)+g_{kl}\left(\beta_{kl}^{(1)}\right)+\int_{\beta_{kl}^{(1)}}^{\beta_{kl}^{(2)}}\left|g_{kl}^{\prime}(v)\right|dv\right).
\end{equation}
 It is worth mentioning that $q_{kl}$ and hence $g_{kl}$ are differentiable
over their respective intervals as the sign of $S^{\prime\prime}(x_{k}(u))S^{\prime\prime}(x_{l}(u))$
does not change over the interval $[u_{0},u_{0}+\alpha]$. We then
have $\left|I_{4}(k,l)\right|\leq\tau \mathcal{H}$ where $\mathcal{H}$ is some constant
independent of $\tau$. Hence, $\lim_{\tau\to0}I_{4}(k,l)=0,\forall k,l$.
By a similar argument, $\lim_{\tau\to0}I_{5}(k,l)=0,\forall k,l$.
From Equations~\ref{eq:I2} and \ref{eq:I3equality}, we get $I_{2}=0$.
Since 
\begin{equation}
\lim_{\alpha\to0}\frac{1}{\alpha}I_{1}=\frac{1}{L}\sum_{k=1}^{N(u_{0})}\frac{1}{\left|S^{\prime\prime}(x_{k})\right|}=P(u_{0}),\label{eq:I1equalsP}
\end{equation}
 the main result stated in Theorem~\ref{GradDensityID} follows.
\qed

\section{Experimental verification}
Below, we show comparisons between our wave function method
and the characteristic function approach on some trigonometric and
exponential functions sampled on a regular grid between $[-0.125,0.125]$
at a grid spacing of $\frac{1}{2^{15}}$. For the sake of convenience,
we normalized the functions such that its maximum gradient value is
1. Using the sampled values $\hat{S}$, we computed the Fast Fourier
transform of $\exp\left(\frac{i\hat{S}}{\tau}\right)$ at $\tau=0.00001$,
took its magnitude squared and then normalized it to obtain the density function of
its derivative. We also computed the discrete derivative of $S$ ($\hat{s}$) at these grid
locations and then obtained the density from the inverse Fourier transform of the characteristic function
of the random variable $s(X)$. The plots shown in Figure~\ref{fig:HistComparisons}
provide anecdotal empirical evidence, corroborating our mathematical result
stated in Theorem~\ref{GradDensityID}. Notice the near-perfect
match between the gradient densities computed via the characteristic function approach
and the gradient densities determined using our wave function method.
\begin{center}
\begin{figure}[ht]
\begin{centering}
\begin{minipage}[c]{0.5\linewidth}%
\includegraphics[width=0.97\textwidth]{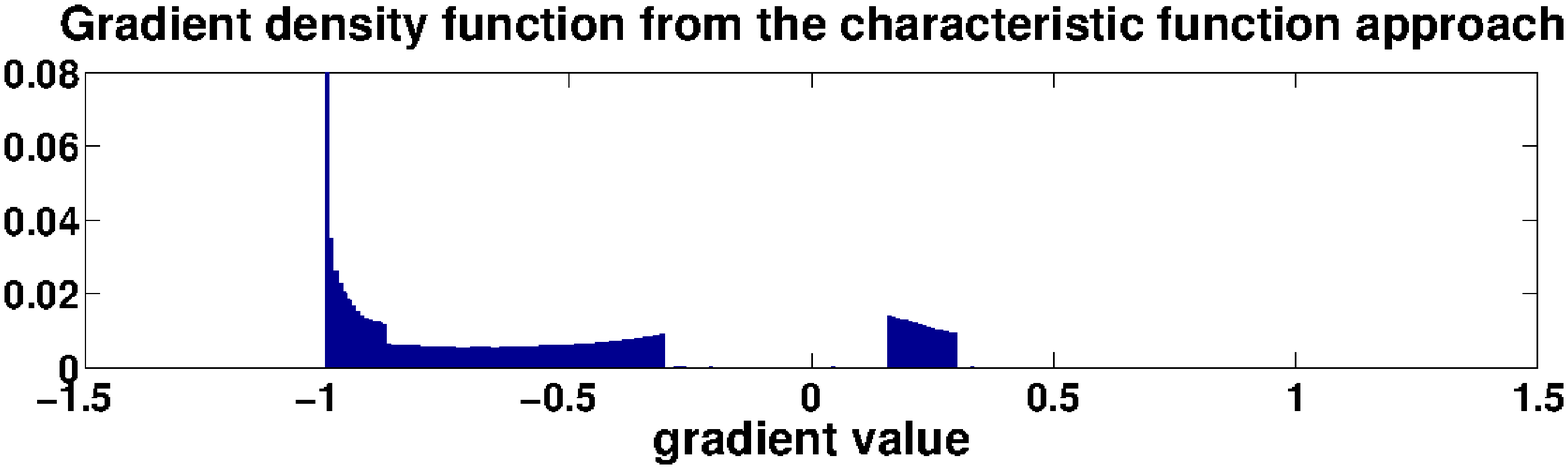} %
\end{minipage}%
\begin{minipage}[c]{0.5\linewidth}%
\includegraphics[width=0.97\textwidth]{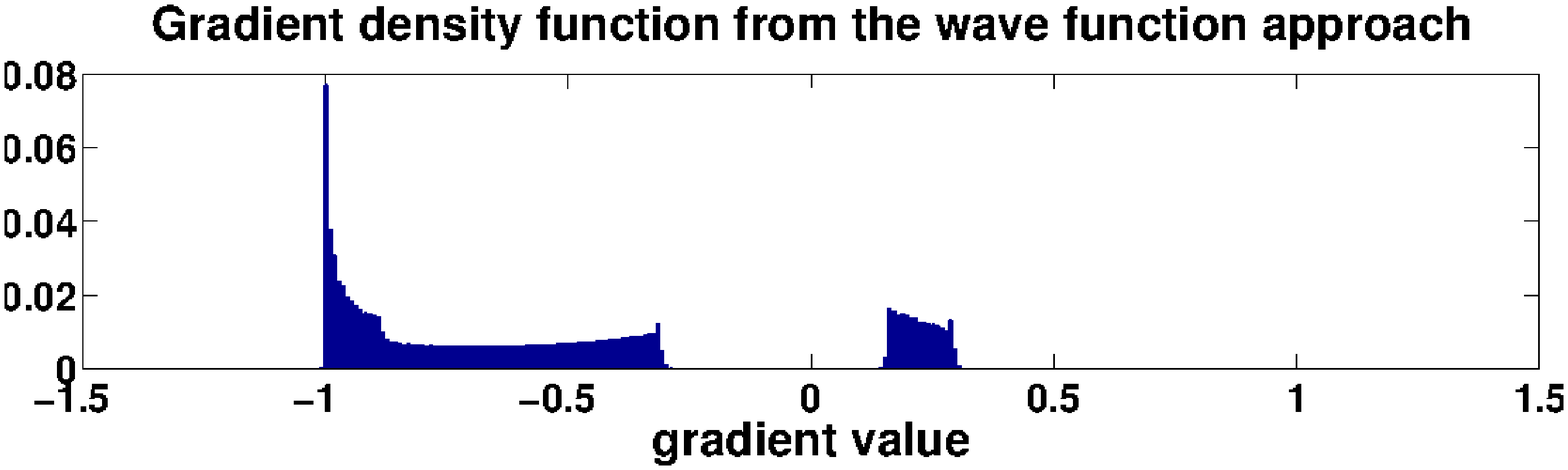} %
\end{minipage}\\
\begin{minipage}[c]{0.5\linewidth}%
\includegraphics[width=0.97\textwidth]{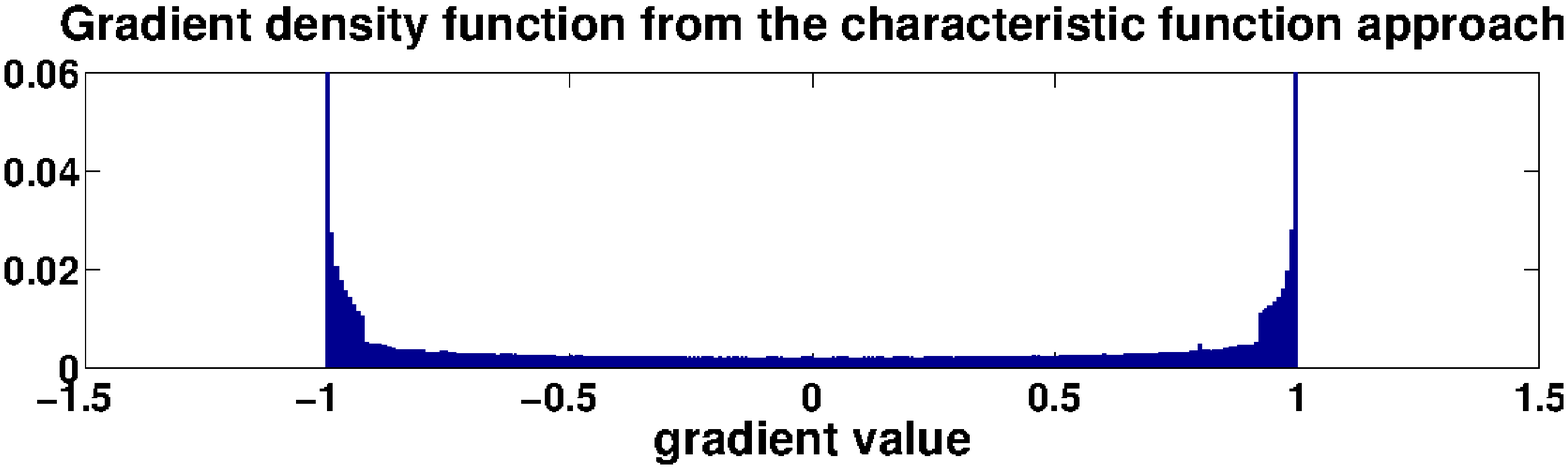} %
\end{minipage}%
\begin{minipage}[c]{0.5\linewidth}%
\includegraphics[width=0.97\textwidth]{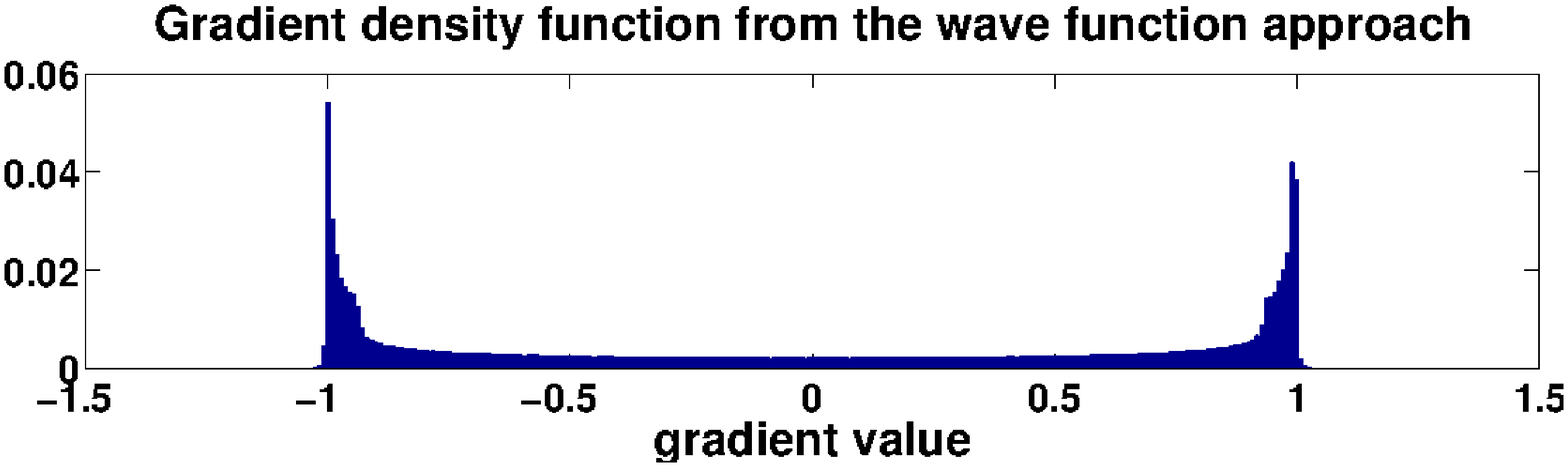} %
\end{minipage}\\
\par\end{centering}
\caption{Comparison results. (i) Left: Gradient densities obtained from the characteristic function,
(ii) Right: Gradient densities obtained from the wave function.}
\label{fig:HistComparisons} %
\end{figure}
\par\end{center}
In the middle four plots of Figure~\ref{fig:ConvergenceResults}, we visualize the density function computed using our wave function method for different values of $\tau \in \{0.0003, 0.0001, 0.00005, 0.00001\}$ (see left to right, bottom to top). Observe that as $\tau$ is decreased, our density function steadily approaches the density function obtained via the characteristic function formulation displayed in the top part of Figure~\ref{fig:ConvergenceResults}. We also plot the $\ell_1$ error between these two approaches for various $\tau$ values in the bottom part of Figure~\ref{fig:ConvergenceResults}. The $\ell_1$ error $\mathcal{E}$ is defined as
\begin{equation}
\mathcal{E} \equiv \sum_{i=1}^{\# bins} \left | \hat{P_{\tau}}_{i} - \hat{P}_{i} \right |,
\end{equation}
where $\hat{P_\tau}_{i}$ and $\hat{P}_{i}$ are the estimated densities of our wave function method and the characteristic function approach respectively, at the $i^{th}$ frequency bin.
As we would expect, the $\ell_1$ error converges towards zero as $\tau$ is decreased.
\begin{center}
\begin{figure}[ht]
\begin{centering}
\begin{minipage}[c]{0.5\linewidth}
\includegraphics[width=1\textwidth]{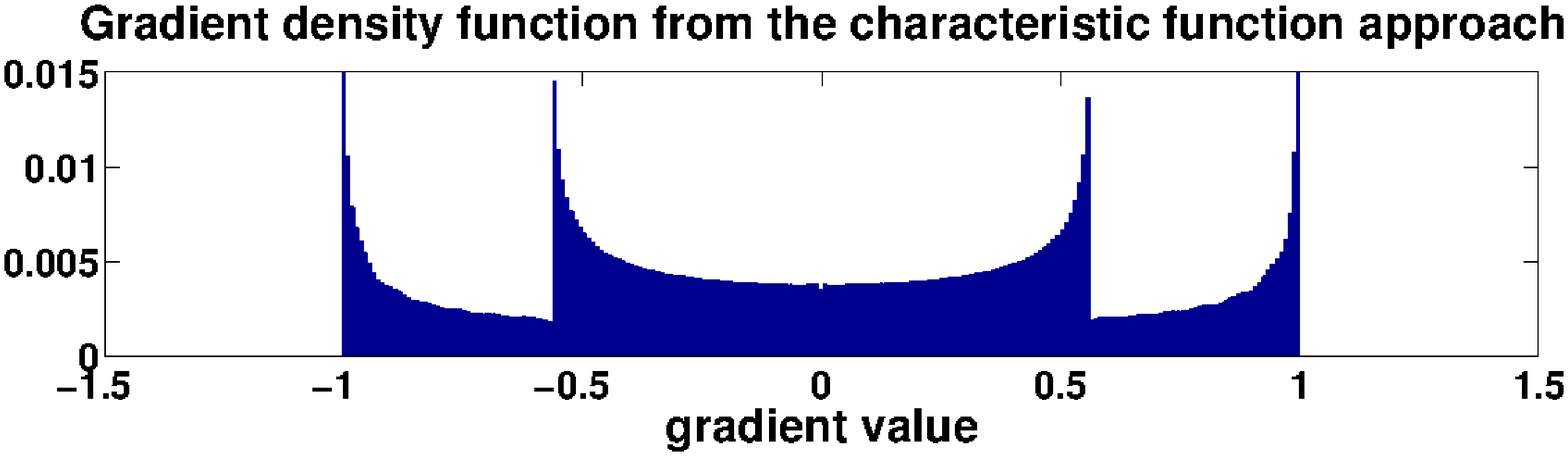} 
\end{minipage}
\par\end{centering}
\begin{centering}
\begin{minipage}[c]{0.5\linewidth}%
\includegraphics[width=0.97\textwidth]{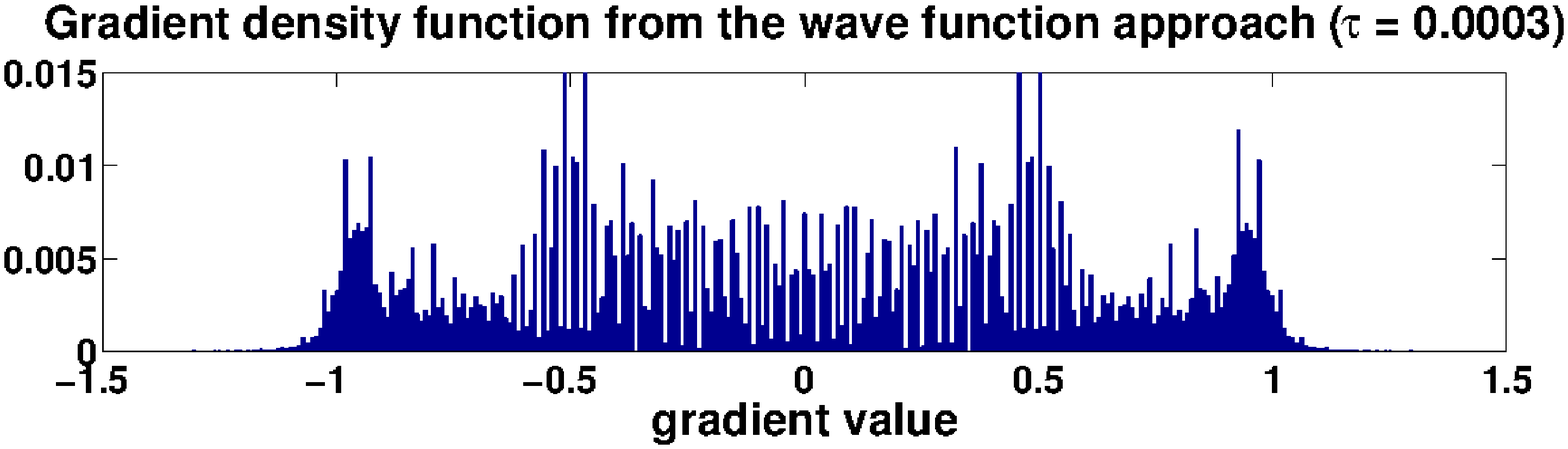} %
\end{minipage}%
\begin{minipage}[c]{0.5\linewidth}%
\includegraphics[width=0.97\textwidth]{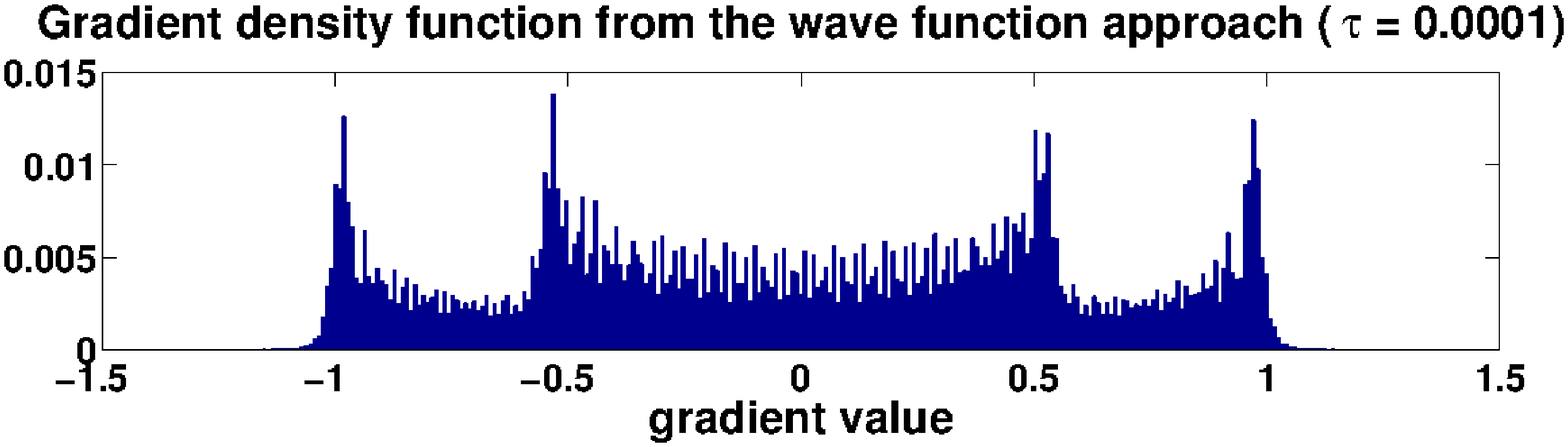} %
\end{minipage}
\begin{minipage}[c]{0.5\linewidth}%
\includegraphics[width=0.97\textwidth]{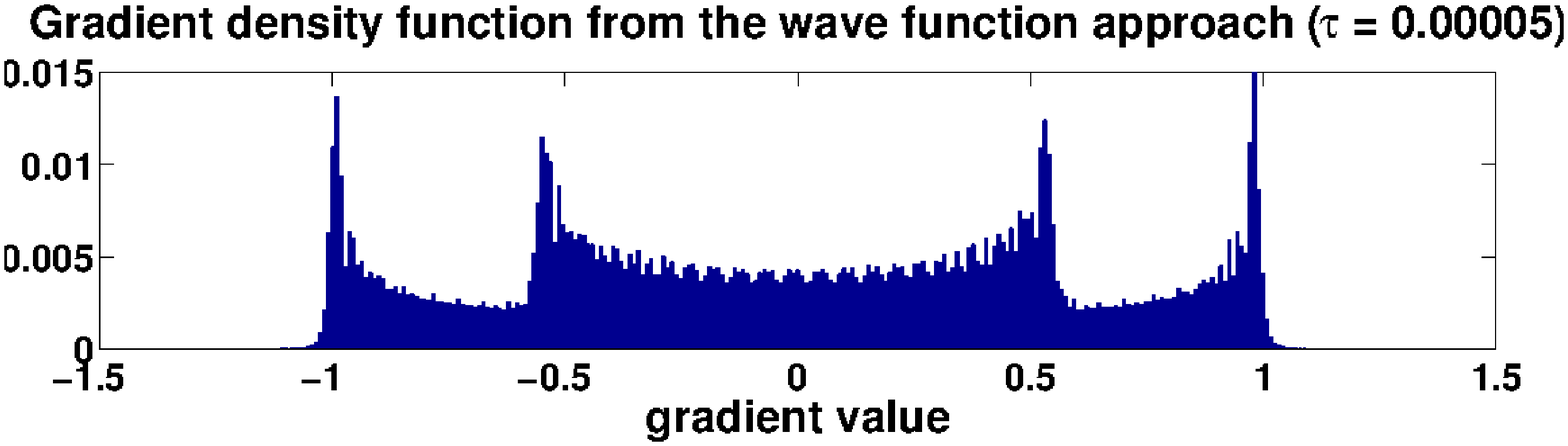} %
\end{minipage}%
\begin{minipage}[c]{0.5\linewidth}%
\includegraphics[width=0.97\textwidth]{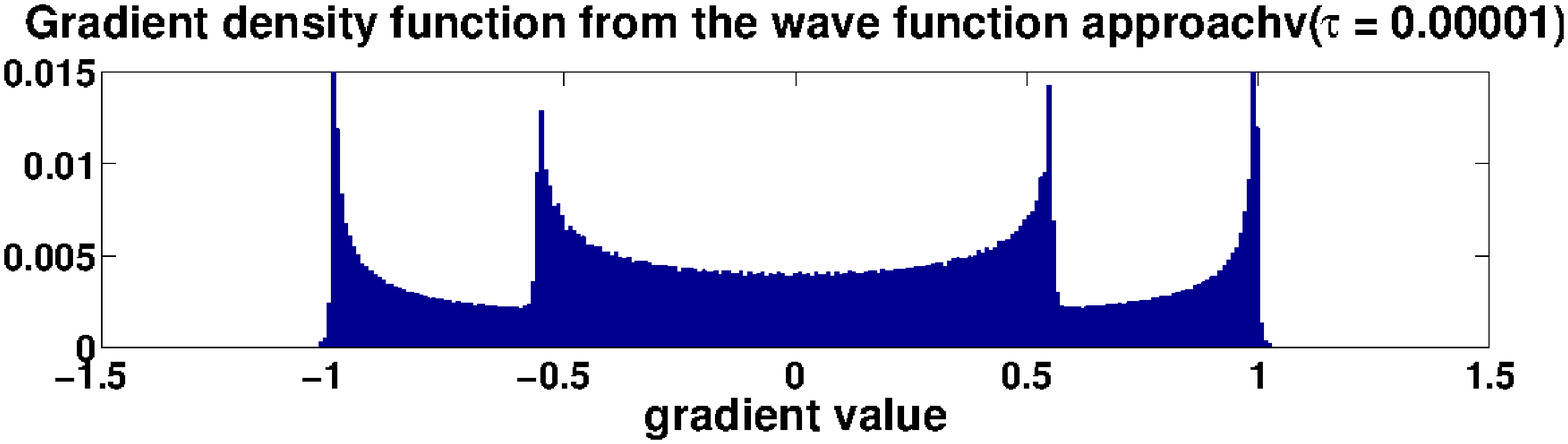} %
\end{minipage}
\par\end{centering}
\begin{centering}
\begin{minipage}[c]{0.5\linewidth}
\includegraphics[width=1\textwidth]{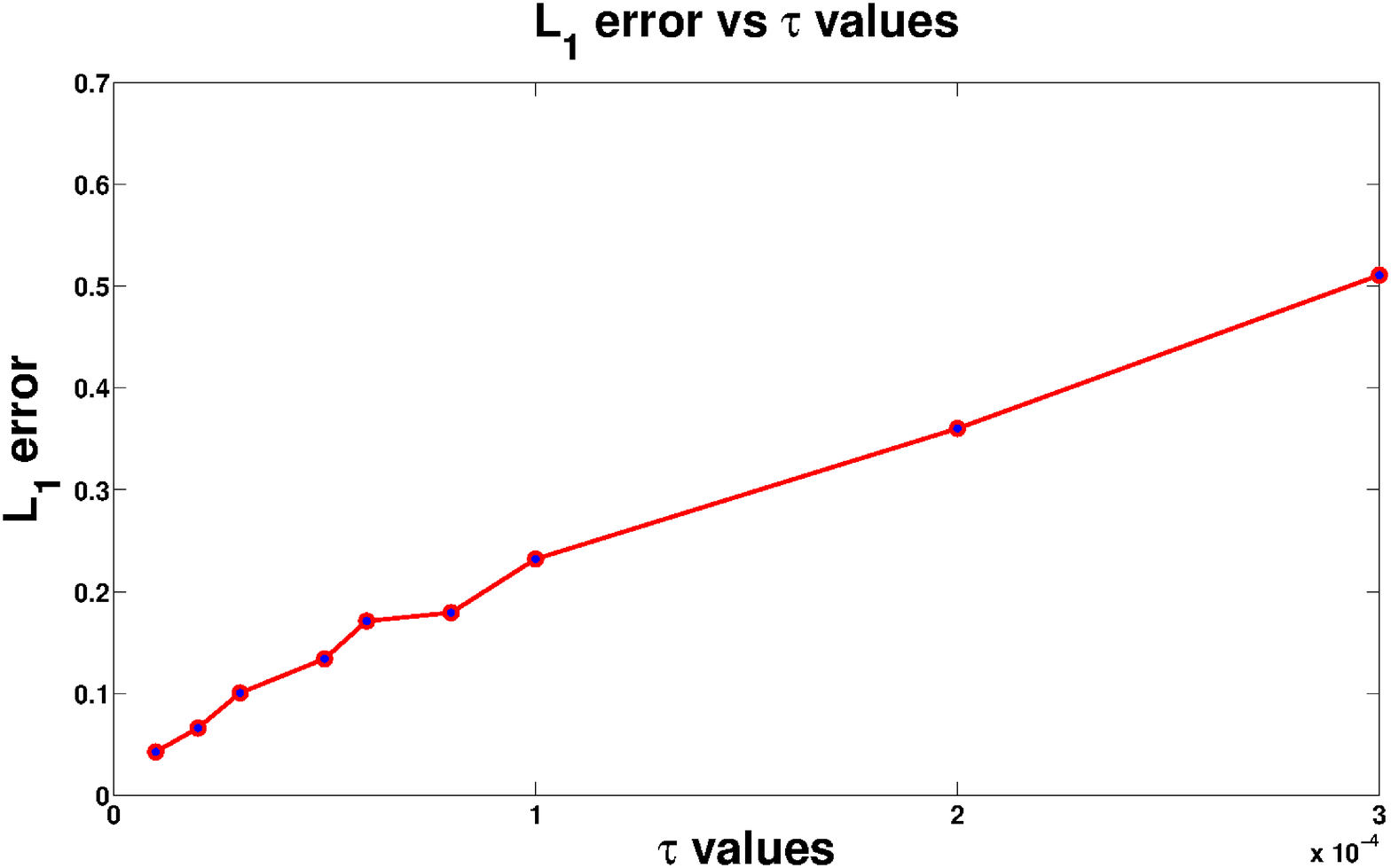} 
\end{minipage}
\par\end{centering}
\caption{Convergence results. (i) Top: Gradient density obtained from the characteristic function, \newline 
(ii) Middle: Gradient densities obtained from the wave function for different $\tau$ values, \newline (iii) Bottom: $\ell_1$ error for various $\tau$ values.}
\label{fig:ConvergenceResults} 
\end{figure}
\par\end{center}

\section{Computational complexity}
Though not central to the main theme of our current work, we would like to add the following note on the computational time complexity of our wave function method in comparison to the characteristic function formulation. Given the $N$ sampled values $\hat{S}$ and its derivative $\hat{s}$, the characteristic function defined in Equation~\ref{eq:characfunc} needs to be computed for $N$ integral values of $\omega$ ranging from $-\frac{N-1}{2}$ to $\frac{N-1}{2}$ if $N$ is odd and from $-\frac{N}{2}$ to $\frac{N}{2}$ if $N$ is even. Each value of $\omega$ requires summation over the $N$ sampled values of $\exp\left(i\omega s(x)\right)$. Hence the total time required to determine the characteristic function is $O(N^2)$.  The density function of the derivative is obtained via the inverse Fourier transform of the the characteristic function, which is an $O(N \log N)$ operation \cite{Cooley65}. The overall time complexity is therefore $O(N^2)$. For our wave function method, the Fourier transform of $\exp\left(\frac{i\hat{S}(x)}{\tau}\right)$ can be computed in $O(N \log N)$ \cite{Cooley65} and the subsequent squaring operation to obtain the power spectrum can be performed in $O(N)$. Hence the density function can be determined in $O(N \log N)$. Our method compares favorably with respect to the characteristic function approach in the overall time complexity.

\section{Discussion}

Observe that the integrals 
\begin{equation}
I_{\tau}(u_{0})=\int_{u_{0}}^{u_{0}+\alpha}P_{\tau}(u)du,\hspace{10pt}I(u_{0})=\int_{u_{0}}^{u_{0}+\alpha}P(u)du
\end{equation}
 give the interval measures of the density functions $P_{\tau}$ and
$P$ respectively. Theorem~\ref{GradDensityID} states that at
small values of $\tau$, both the interval measures are approximately
equal, with the difference between them being $o(1)$. Recall that
by definition, $P_{\tau}$ is the normalized power spectrum \cite{Bracewell99} of the wave function $\phi(x)=\exp\left(\frac{iS(x)}{\tau}\right)$.
Hence we conclude that the power spectrum of $\phi(x)$ can potentially
serve as a \emph{density estimator} for the gradients of $S$ at small
values of $\tau$. The experimental results shown above serve as a
demonstration, anecdotally attesting to the verity of the result.  We also built an informal bridge between our wave function method 
and the characteristic function approach for estimating probability densities by directly trying to recast the former expression into the latter. The difficulties faced in relating the two approaches made the stationary phase approximation method an inevitable tool to formally prove Theorem~\ref{GradDensityID}. 

Our result is directly inspired by the three-way relationships between
the classical momentum $\nabla S$, the quantum momentum operator
$-i\hbar\frac{\partial}{\partial x}$ and its spatial frequency spectrum.
Since these relationships hold in higher dimensions as well, we are
likewise interested in extending our density estimation result to
higher dimensions. This is a fruitful avenue for future work.

\bibliographystyle{elsarticle-harv}
\bibliography{DensityEstimation}

\appendix
\section{Proof of Lemmas}
\label{sec:Proof-of-Lemmas}

\noindent \emph{1. Proof of Finiteness Lemma}\\
 We prove the result by contradiction. Observe that $\mathcal{A}_{u}$
is a subset of the compact set $\Omega$. If $\mathcal{A}_{u}$ is
not finite, then by Theorem~2.37 in \cite{Rudin76}, $\mathcal{A}_{u}$
has a limit point $x_{0}\in\Omega$. Consider a sequence $\{x_{n}\}_{n=1}^{\infty}$,
with each $x_{n}\in\mathcal{A}_{u}$, converging to $x_{0}$. Since
$s(x_{n})=u,\forall n$, from the continuity of $s$
we get $s(x_{0})=u$ and hence $x_{0}\in\mathcal{A}_{u}$. If $x_0 = b_1$ or $x_0=b_2$,
then $u\in\mathcal{C}$ giving us a contradiction. Otherwise we get
\begin{equation}
\lim_{n\rightarrow\infty}\frac{s(x_{0})-s(x_{n})}{x_{0}-x_{n}}=0=s^{\prime}(x_0)=S^{\prime\prime}(x_{0}).
\end{equation}
 Based on the definitions given in Equation~\ref{AandCset}, we have $x_{0}\in\mathcal{B}$
and $u\in\mathcal{C}$ again resulting in a contradiction. \\
\qed

\noindent \emph{2. Proof of Interval Lemma}\\
 Observe that $\mathcal{B}$ is closed because if $x_{0}$ is a
limit point of $\mathcal{B}$, from the continuity of $S^{\prime\prime}$
we have $S^{\prime\prime}(x_{0})=0$ and hence $x_{0}\in\mathcal{B}$.
$\mathcal{B}$ is also compact as it is a closed subset of $\Omega$.
Since $s$ is continuous, $\mathcal{C}=s(\mathcal{B})\cup\{s(b_{1}),s(b_{2})\}$
is also compact and hence $\mathbb{R}-\mathcal{C}$ is open. Then
for $u\notin\mathcal{C}$, there exists an open neighborhood $\mathcal{N}_{r}(u)$
for some $r>0$ around $u$ such that $\mathcal{N}_{r}(u)\cap\mathcal{C}=\emptyset$.
By defining $\eta=\frac{r}{2}$, the proof is complete.  \\
\qed

\noindent \emph{3. Proof of Density Lemma}\\
 Since the random variable $X$ is assumed to have a uniform distribution
on $\Omega$, its density is given by $f_{X}(x)=\frac{1}{L}$ for
every $x\in\Omega$. Recall that the random variable $Y$ is obtained
via a random variable transformation from $X$, using the function
$s$. Hence, its density function exists on $\mathbb{R}-\mathcal{C}$---
where we have banished the image (under $s$) of the measure
zero set of points where $S^{\prime\prime}$ vanishes---and is given
by Equation~\ref{eq:graddensity}. The reader may refer to \cite{Billingsley95}
for a detailed explanation.  \\
\qed 
\end{document}